\documentclass[sigconf]{acmart}
\AtBeginDocument{%
  }

\setcopyright{acmlicensed}
\copyrightyear{2026}
\acmYear{2026}
\acmDOI{XXXXXXX.XXXXXXX}
\acmISBN{978-1-4503-XXXX-X/2018/06}

\usepackage{graphicx}
\usepackage{booktabs}
\usepackage{caption}
\usepackage{colortbl}
\usepackage{ulem}  
\usepackage{array}
\usepackage{multirow}
\usepackage{xcolor}            
\usepackage{colortbl}          
\begin{document}

\title{H4G: Unlocking Faithful Inference for Zero-Shot Graph Learning in Hyperbolic Space}

\author{Heng Zhang}
\affiliation{
  \institution{South China Normal University}
  \country{China}}
\email{2024025450@m.scnu.edu.cn}

\author{Tianyi Zhang}
\affiliation{
  \institution{Uber Technologies Inc.}
  \country{USA}}
\email{tianyizhg@gmail.com}

\author{Zijun Liu}
\affiliation{
  \institution{iAUTO}
  \country{China}}
\email{zl3031@columbia.edu}

\author{Yuling Shi}
\affiliation{
  \institution{Shanghai Jiao Tong University}
  \country{China}}
\email{yuling.shi@sjtu.edu.cn}

\author{Yaomin Shen}
\affiliation{
  \institution{Nanchang Research Institute, Zhejiang University}
  \country{China}}
\email{coolshennf@gmail.com}

\author{Haochen You}
\affiliation{
  \institution{Columbia University}
  \country{USA}}
\email{hy2854@columbia.edu}

\author{Haichuan Hu}
\affiliation{
  \institution{Alibaba Cloud}
  \country{China}}
\email{huhaichuan.hhc@alibaba-inc.com}

\author{Lubin Gan}
\affiliation{
  \institution{University of Science and Technology of China}
  \country{China}}
\email{ganlubin@mail.ustc.edu.cn}

\author{Jin Huang}
\authornote{Corresponding author.}
\affiliation{
  \institution{South China Normal University}
  \country{China}}
\email{huangjin@m.scnu.edu.cn}

\renewcommand{\shortauthors}{Trovato et al.}

\begin{abstract}
Text-attributed graphs are widely used across domains, offering rich opportunities for zero-shot learning via graph-text alignment. However, existing methods struggle with tasks requiring fine-grained pattern recognition, particularly on heterophilic graphs. Through empirical and theoretical analysis, we identify an \textbf{over-abstraction problem}: current approaches operate at excessively large hyperbolic radii, compressing multi-scale structural information into uniform high-level abstractions. This abstraction-induced information loss obscures critical local patterns essential for accurate predictions. By analyzing embeddings in hyperbolic space, we demonstrate that optimal graph learning requires \textbf{faithful preservation} of fine-grained structural details, better retained by representations positioned closer to the origin. To address this, we propose \textbf{H4G}, a framework that systematically reduces embedding radii using learnable block-diagonal scaling matrices and Möbius matrix multiplication. This approach restores access to fine-grained patterns while maintaining global receptive ability with minimal computational overhead. Experiments show H4G achieves state-of-the-art zero-shot performance with \textbf{12.8\%} improvement on heterophilic graphs and \textbf{8.4\%} on homophilic graphs, confirming that radius reduction enables faithful multi-scale representation for advancing zero-shot graph learning.
\end{abstract}



\maketitle

\section{Introduction}
Zero-shot graph learning focuses on building models that can accurately predict on text-attributed graphs without relying on labeled training data from the target domain \cite{GraphText,graphprompt,graphclip}. Text-attributed graphs, widely used in areas such as social networks, academic citations \cite{graphinsight}, e-commerce systems \cite{GraphInstruct}, and biology\cite{eddy2024graph}, combine textual content at each node with graph structures. This combination naturally enables knowledge transfer across different domains. Traditional graph learning methods often struggle to make full use of the rich semantic information embedded in textual descriptions \cite{guo2024zeroshot}. By aligning graph representations with textual embeddings, zero-shot learning methods can leverage advancements in language models to navigate unseen domains and tasks. Recent breakthroughs in large language models, showcasing their cross-domain adaptability, have further inspired efforts to integrate textual and structural features \cite{ramasesh2024fine}.

\begin{figure}[tb!]
  \vspace{0.10in}
    \centering
    \includegraphics[width=\columnwidth]{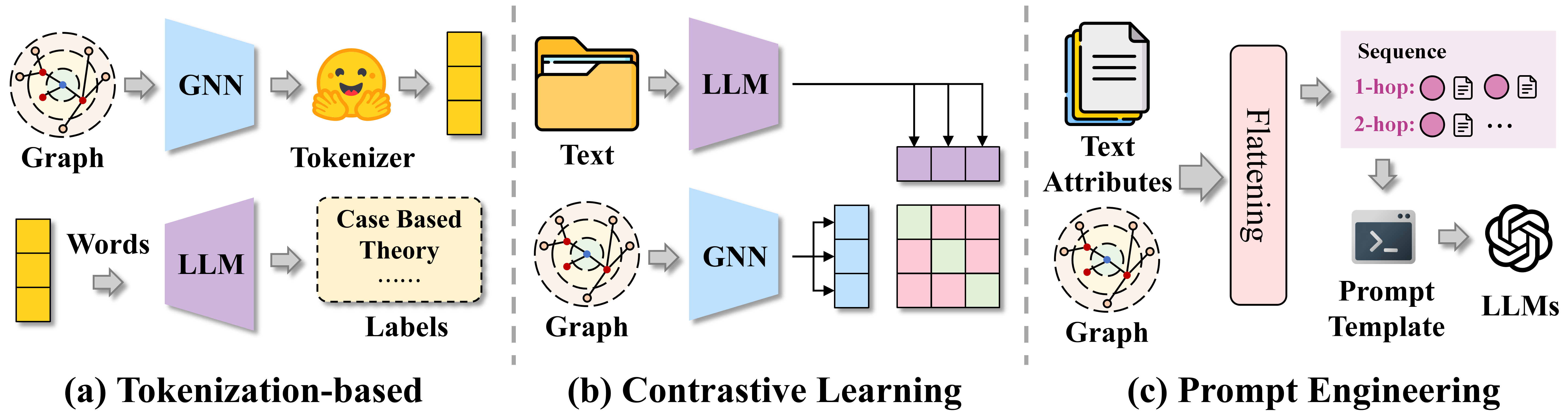}
    \caption{ Three main categories for Integrating Graphs with Large Language Models. 
    }
    \label{motivation}

\end{figure}

Recent progress in zero-shot graph learning primarily follows three key paradigms for graph-text alignment. The first paradigm leverages large language models as enhancers or predictors by converting graph structures into textual sequences compatible with LLMs \cite{li2024urbangpt,he2023harnessing,zhao2024gimlet}. GraphGPT \cite{graphgpt} and LLaGA \cite{chen2024llaga} exemplify this method through projection layers that map graph representations into LLM token spaces. While achieving reasonable results, these methods often suffer information loss during graph-to-text conversion and demand significant computational resources. The second paradigm focuses on contrastive learning to align graph and text modalities \cite{congrat,unigraph,natureisneed}. GraphCLIP \cite{graphclip} notably uses LLMs to generate graph-summary pairs and trains contrastive objectives to align graph encoders with textual representations. This approach demonstrates strong performance in cross-domain tasks by explicitly bridging graph features with text embeddings. The third paradigm pursues direct alignment between graph and text representations through pretraining GNNs to match frozen LLM token embeddings \cite{ENGINE,GraphLLM}. TEA-GLM \cite{wang2024llms} represents this line of work and offers promising results in zero-shot scenarios. However, current methods in this category often assume fixed alignment granularity, limiting their adaptability to diverse datasets and tasks.

We analyzed the learned representations of current graph-text alignment methods across diverse graph learning scenarios. We used hyperbolic geometry as our analytical framework. Leading approaches including GraphCLIP \cite{graphclip}, ZeroG \cite{zerog}, and OFA \cite{OFA} were evaluated on various graph datasets. Their embeddings consistently positioned far from the hyperbolic origin, typically at radii around 7-8 (Figure 2). This positioning reveals a critical issue. In hyperbolic space, radius directly encodes abstraction level. Points near the origin preserve fine-grained details. Distant points compress information into coarse summaries. The large radii indicate that existing methods operate at excessive abstraction levels. This over-abstraction creates systematic performance problems. In node classification tasks, the compressed representations fail to distinguish structurally similar but semantically different nodes. Tasks requiring fine-grained structural discrimination also struggle. Critical local patterns become obscured by high-level abstractions. The problem intensifies in heterophilic graphs. Local structural nuances are essential for accurate predictions in these graphs. Our analysis exposes a core limitation: current alignment methods sacrifice the multi-scale structural information needed to differentiate semantically distinct subgraphs. This mirrors the over-globalizing issue in Graph Transformers, but manifests differently. There, excessive attention to distant nodes weakens local information. Here, excessive abstraction obscures the fine-grained patterns necessary for effective graph-text alignment.

\begin{figure}[tb!]
    \centering
    \includegraphics[width=0.97\columnwidth]{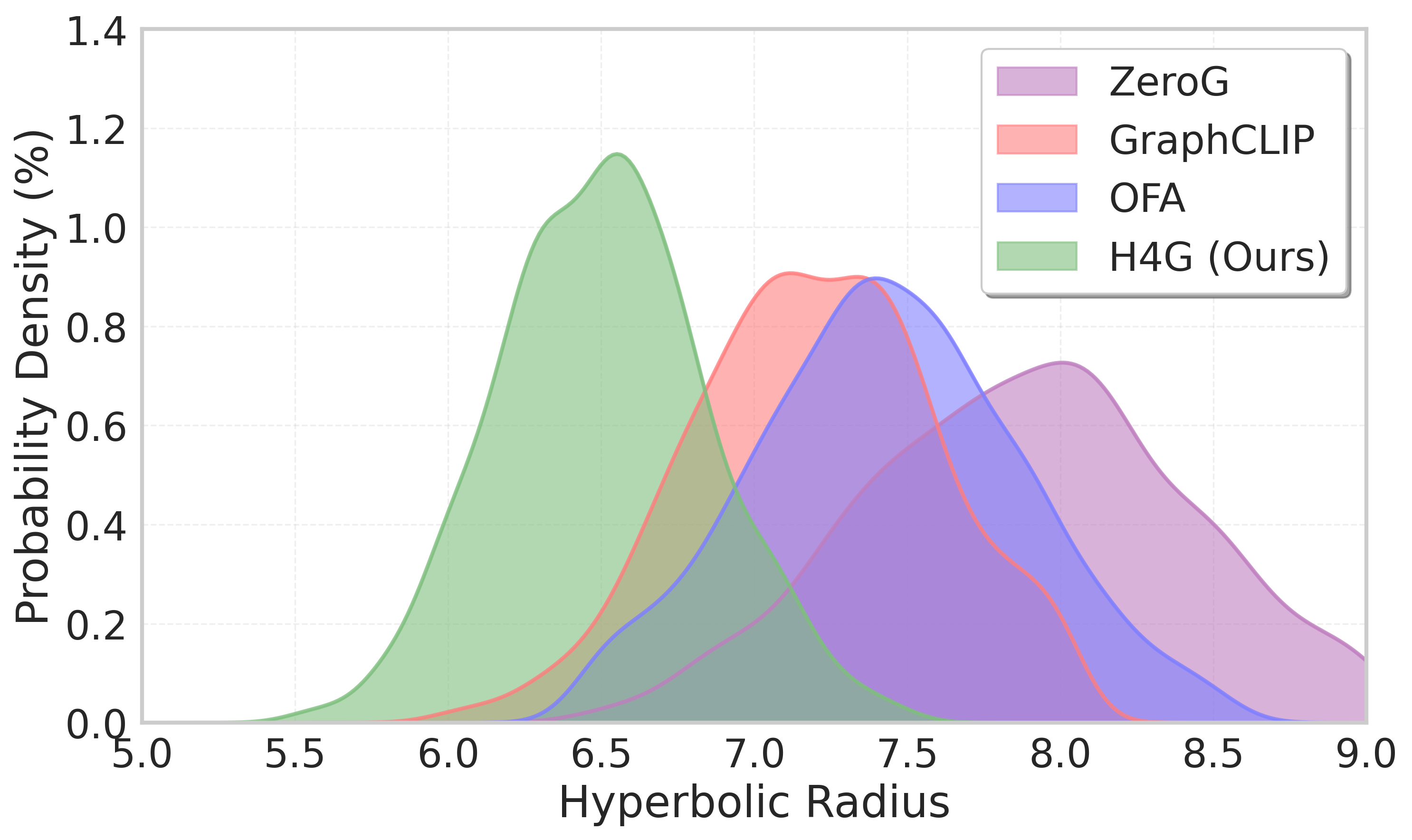}
    \caption{Hyperbolic radius distribution of text embeddings. Baseline methods operate at large radii with dispersed distributions, losing fine-grained structural information. H4G achieves systematic radius reduction with concentrated embeddings, preserving detailed graph patterns essential for effective zero-shot learning.
    }
    \label{figure2}
    \vspace{-0.15in}
\end{figure}

This pattern points to a deeper consideration about how graph-text alignment preserves the multi-scale nature of graph structures. Graph neural networks naturally build hierarchical representations through message passing. Initial layers encode fine-grained neighborhood patterns. Subsequent layers capture broader structural contexts. This layered processing creates a natural hierarchy of information granularity. Current alignment methods undermine this multi-scale nature. They position these rich representations at large hyperbolic radii. The result is compression of hierarchical information into a single abstraction level. This reduces the expressiveness of graph representations. It also limits their ability to convey nuanced structural relationships from the original data. The information loss proves particularly detrimental because graph learning tasks often require reasoning across multiple structural scales simultaneously. Deep GNNs suffer from over-smoothing that prevents access to fine-grained distinctions. Alignment methods operating uniformly at high abstraction levels face a parallel problem. Both prevent models from achieving faithful representation. Preserving structural details appears especially important for graph learning. Even seemingly global tasks often rely on local discriminative patterns for differentiation. Moving embeddings closer to the hyperbolic origin could restore access to these multi-scale features. The alignment would then better preserve the inherent complexity of graph structures. This suggests a clear path forward. Systematically reducing the hyperbolic radius might enable more complete utilization of the structural information encoded by graph neural networks.

To mitigate over-abstraction and retain global receptive ability, we propose H4G, a novel framework designed to optimize embedding radius in graph learning. H4G systematically reduces hyperbolic radii of embeddings, recovering access to fine-grained structural information lost in conventional alignment methods. Our approach introduces learnable block-diagonal scaling matrices operating through Möbius matrix multiplication. Embeddings progressively shift closer to the hyperbolic origin. This radius reduction ensures representations retain hierarchical and multi-scale features critical for accurate graph learning, avoiding abstraction-induced information loss observed in previous methods. Previous methods collapse hierarchical information into uniform high-level abstractions. H4G enables faithful multi-scale representation by explicitly controlling abstraction level through radius adjustment. By dynamically learning optimal scaling factors for radius reduction, H4G adapts representational granularity to structural details inherent in various graph tasks. This systematic reduction of abstraction levels creates a unified solution for preserving crucial local patterns, mitigating limitations of fixed-abstraction paradigms. H4G achieves this with minimal computational overhead, offering a lightweight yet effective mechanism to enhance graph-text alignment and improve zero-shot inference across diverse graph learning scenarios. Our contributions can be summarized as follows:
\begin{itemize}
\item We identify the over-abstraction problem in graph-text alignment, showing that large hyperbolic radii compress multi-scale information and obscure fine-grained patterns.
\item We propose \textbf{H4G}, a framework that reduces embedding radii to restore fine-grained structural information while maintaining global receptive ability.
\item We design learnable block-diagonal scaling matrices using Möbius multiplication to achieve efficient radius reduction with explicit granularity control.
\item Experiments show H4G improves zero-shot accuracy by \textbf{12.8\%} on heterophilic graphs and \textbf{8.4\%} on homophilic graphs, outperforming state-of-the-art methods.
\end{itemize}

\begin{figure*}[t]
   \centering
   \includegraphics[width=\textwidth]{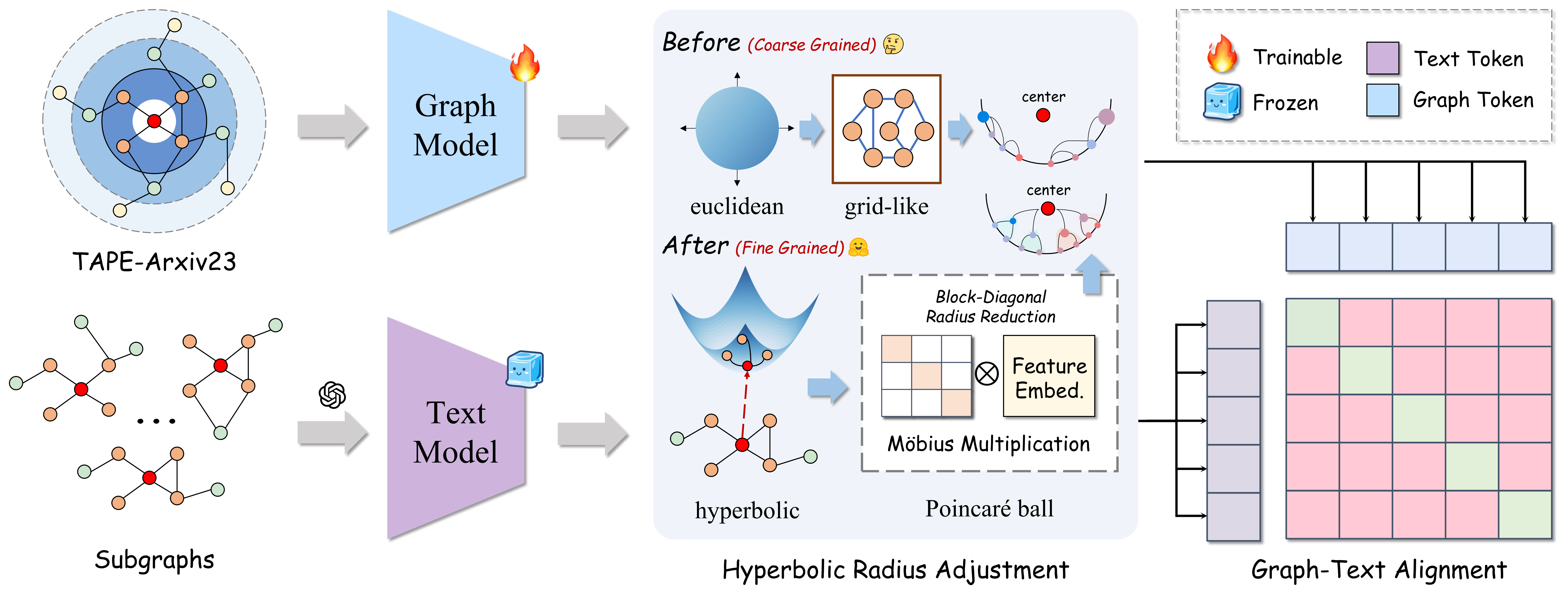}
\caption{Overview of the H4G Framework. H4G achieves fine-grained representation learning through radius adjustment in hyperbolic space. Text-attributed graphs are encoded by graph and text models and then projected into the Poincaré ball of hyperbolic space. H4G employs learnable block-diagonal scaling matrices with Möbius matrix multiplication to systematically adjust embeddings from high abstraction levels (far from origin) to fine-grained information levels (close to origin), preserving richer structural details in the representation space for high-quality graph-text alignment in zero-shot learning.}
   \label{fig:main}
\end{figure*}

\section{Related Work}
\subsection{Zero-shot and Few-shot Learning on Graphs}
Real-world graph applications often encounter the challenge of limited labeled data, making few-shot and zero-shot learning essential for practical deployment~\cite{liu2025imbalanced,chen2024taskequivariant}. Early approaches adopt meta-learning paradigms to achieve rapid adaptation using minimal examples~\cite{liu2024learningnovel,chen2024metadiff}. Self-supervised methods, such as DGI and GraphCL, enhance node representations through contrastive pre-training~\cite{yan2025selfcumulative,li2024selfsupervisedgraph}. Models like MVGRL incorporate subgraph and diffusion information to capture richer graph semantics. However, these methods typically rely on task-specific fine-tuning and suffer substantial performance drops when supervisory signals are extremely sparse~\cite{jung2024simple,wang2025generalized}. Recent efforts have explored pseudo-labeling techniques that expand labeled data by generating confident predictions for unlabeled nodes~\cite{li2024pseudocontrastive,ding2024dataefficientgraph}. The rise of large language models (LLMs) offers promising new solutions to these limitations~\cite{chen2024exploringpotential,han2024largelanguagemodels}. For instance, encoding strategies in LLaGA enable LLMs to process graph data directly without additional training~\cite{chen2024llaga,he2024modelgeneralization}. TEA-GLM aligns GNN representations with LLM embeddings to achieve strong generalization across tasks and datasets~\cite{fan2024graphmachinelearning,chen2024llaga}. These advancements highlight how LLMs can address graph learning challenges without the need for task-specific supervision. Distinct from prior studies, our proposed method introduces a novel framework that systematically reduces reliance on labeled data while preserving fine-grained structural information critical for efficient zero-shot learning~\cite{ma2024graphcontrastive,yu2024surveyfewshot}.

\subsection{Learning in Hyperbolic Space}
Unlike Euclidean geometry, hyperbolic space functions as the continuous analog of a tree, making it particularly suitable for modeling the hierarchical structures embedded in complex systems~\cite{yang2024hyperbolicgraphneural,li2024hyperbolicgraphneuraltemporal}. Prior studies have successfully applied hyperbolic representations across various domains~\cite{zheng2025hyperbolicgraphwavelet,chen2025hyperbolicgenomeembeddings}. For example, molecular structures benefit from encoding chemical hierarchies~\cite{lin2025treewassersteindistancehigh,grover2025spectroriemanniangraphneural}. 3D data embedding leverages spatial hierarchies~\cite{yang2024hyperbolicgraphneuraltutorial,zheng2025hyperbolicgraphwavelet}. Text data preserves semantic and taxonomic relationships~\cite{shin2024learningstructuredrepresentations,ye2025hyperbolicbernsteinneural}. Image data captures rich structural and semantic hierarchies. These results highlight how hyperbolic embeddings adapt naturally to hierarchical data, offering compact and expressive representations for diverse tasks~\cite{li2025dhhnn,wang2024geometryawarealgorithm}. However, most existing research primarily relies on the tree-like geometric properties of hyperbolic space, without examining the role of hyperbolic radius in reflecting nuanced semantic or structural variations~\cite{chen2024curvaturelearninggeneralization,ayoughi2025designinghierarchiesoptimal}. In our work, we focus on this underexplored dimension by systematically investigating how the hyperbolic radius encodes detailed structural granularity and finer semantic distinctions~\cite{chen2025gyrogroupbatchnormalization,shin2024learningstructuredrepresentations}. This perspective advances graph-text alignment by promoting a more faithful retention of structural details, mitigating critical challenges in fine-grained pattern recognition and enabling reliable zero-shot learning across complex graph tasks~\cite{yang2024hyperbolicgraphneural,li2025dhhnn}.

\section{Preliminaries}
\subsection{Hyperbolic Geometry}
We adopt the Poincaré ball model $\mathcal{D}^d = \{x \in \mathbb{R}^d : \|x\| < 1\}$ with curvature $-c$ where $c > 0$ for our framework. This model offers a natural way to encode hierarchical information through radial positions. The hyperbolic distance between two points $x, y \in \mathcal{D}^d$ is defined as
\begin{equation}
d_{\mathcal{D}}(x, y) = \frac{2}{\sqrt{c}} \tanh^{-1}(\sqrt{c}\|\ominus_c x \oplus_c y\|)
\end{equation}
where $\oplus_c$ and $\ominus_c$ represent Möbius addition and subtraction operations. The hyperbolic radius $r(x) = d_{\mathcal{D}}(x, 0)$ directly encodes abstraction level. Points near the origin have smaller radii and preserve fine-grained details. In contrast, points farther from the center have larger radii and capture coarse-grained patterns. This radial encoding plays an important role in graph learning. When methods operate at excessively large radii, they compress multi-scale structural information into uniform high-level abstractions. This compression obscures fine-grained discriminative patterns essential for accurate predictions. Understanding this relationship between radius and information granularity motivates our framework design. The exponential map $\exp_0^c$ enables transformation from tangent space to hyperbolic space, facilitating gradient-based optimization while preserving the geometric properties necessary for effective learning.

\subsection{Graph-Text Alignment Framework}
Consider a text-attributed graph $G = (V, E, X)$ where $V$ represents the node set with $|V| = n$ nodes, $E$ denotes the edge set, and $X \in \mathbb{R}^{n \times d_t}$ contains textual features for each node. A graph encoder $f_g: G \rightarrow \mathbb{R}^{d_g}$ produces graph-level representations. A text encoder $f_t: X \rightarrow \mathbb{R}^{d_t}$ generates text embeddings. Traditional alignment methods minimize contrastive losses between corresponding graph-text pairs to learn a shared embedding space. These approaches have achieved success in various graph learning tasks. However, recent analyses reveal an important limitation in these methods. Existing approaches position embeddings at excessively large hyperbolic radii, as we observe in our empirical investigation. This leads to an over-abstraction problem where multi-scale structural information is compressed into uniform high-level representations. The abstraction-induced information loss prevents faithful preservation of fine-grained patterns. These patterns are necessary for distinguishing semantically different but structurally similar subgraphs. The problem becomes particularly evident in heterophilic graphs, where local discriminative features are crucial for accurate predictions. This observation motivates our approach to systematically control and reduce embedding radii.

\section{Methodology}
We propose H4G to mitigate the over-abstraction problem while maintaining global receptive ability. Our framework systematically reduces embedding radii to restore faithful preservation of fine-grained structural information. Through empirical analysis, we observe that existing methods operate at large radii, compressing multi-scale patterns into uniform abstractions. H4G introduces three core components that work together to solve this problem. First, hyperbolic embedding projection maps representations into a space where abstraction levels can be explicitly controlled through radial positions. Second, block-diagonal radius adjustment reduces radii through learnable scaling matrices that operate via Möbius matrix multiplication. Third, hierarchical contrastive learning optimizes the adjusted representations for effective graph-text alignment. These components collectively enable our framework to access fine-grained structural information while maintaining computational efficiency.

\subsection{Hyperbolic Embedding Projection}
Graph learning benefits from representations at multiple abstraction levels simultaneously. Different tasks and graph types require access to both local patterns and global structures. Hyperbolic space provides explicit control over these levels through radial positions, making it well-suited for our framework. We project both graph and text representations into this space to leverage these properties.

Given graph embeddings $\mathbf{h}_g \in \mathbb{R}^{d_g}$ from graph neural networks and text embeddings $\mathbf{h}_t \in \mathbb{R}^{d_t}$ from pretrained language models, we first align their dimensions through linear transformations:
\begin{equation}
\mathbf{h}_g' = \mathbf{W}_g \mathbf{h}_g + \mathbf{b}_g,
\end{equation}
\begin{equation}
\mathbf{h}_t' = \mathbf{W}_t \mathbf{h}_t + \mathbf{b}_t,
\end{equation}
where $\mathbf{W}_g \in \mathbb{R}^{d \times d_g}$ and $\mathbf{W}_t \in \mathbb{R}^{d \times d_t}$ are learnable projection matrices. The bias vectors $\mathbf{b}_g, \mathbf{b}_t \in \mathbb{R}^d$ and unified dimension $d$ ensure both modalities operate in the same space. This dimension alignment is essential for subsequent hyperbolic projection. We then map the transformed representations to the Poincaré ball via exponential maps:
\begin{equation}
\mathbf{z}_g = \exp_0^c(\mathbf{h}_g'), 
\end{equation}
\begin{equation}
  \mathbf{z}_t = \exp_0^c(\mathbf{h}_t')
\end{equation}
The exponential map at the origin with curvature $c > 0$ is defined as:
\begin{equation}
\exp_0^c(\mathbf{v}) = \frac{1}{\sqrt{c}} \tanh(\sqrt{c}\|\mathbf{v}\|) \frac{\mathbf{v}}{\|\mathbf{v}\|}
\end{equation}
This mapping ensures the projected embeddings $\mathbf{z}_g, \mathbf{z}_t \in \mathcal{D}^d$ lie within the Poincaré ball. Their radial positions now directly correspond to abstraction levels, enabling explicit control over information granularity. This controllability forms the foundation for our radius adjustment mechanism.

\subsection{Block-Diagonal Radius Adjustment}
The core innovation of H4G lies in systematically reducing embedding radii to access fine-grained information. We achieve this through learnable block-diagonal scaling matrices applied via Möbius matrix multiplication. This mechanism directly mitigates the over-abstraction problem by bringing representations closer to the origin where structural details can be faithfully preserved. We construct block-diagonal scaling matrices as:
\begin{equation}
\mathbf{S}_g = \text{diag}(\mathbf{S}_{g,1}, \mathbf{S}_{g,2}, \ldots, \mathbf{S}_{g,K}),
\end{equation}
\begin{equation}
\mathbf{S}_t = \text{diag}(\mathbf{S}_{t,1}, \mathbf{S}_{t,2}, \ldots, \mathbf{S}_{t,K}),
\end{equation}
where $\mathbf{S}_{g,k}, \mathbf{S}_{t,k} \in \mathbb{R}^{n \times n}$ represent individual block matrices. The number of blocks $K = d/n$ and block size $n$ control transformation granularity. This block-diagonal structure offers several advantages. It balances parameter efficiency with transformation flexibility, allowing fine-grained control over different embedding dimensions while maintaining computational tractability. The structure also enables independent scaling of different feature subspaces, which proves beneficial for learning task-specific radius adjustments.

The radius-adjusted embeddings are computed through Möbius matrix multiplication:
\begin{equation}
\tilde{\mathbf{z}}_g = \mathbf{S}_g \otimes_c \mathbf{z}_g, 
\end{equation}
\begin{equation}
\tilde{\mathbf{z}}_t = \mathbf{S}_t \otimes_c \mathbf{z}_t
\end{equation}
For a matrix $\mathbf{M} \in \mathbb{R}^{d \times d}$ and vector $\mathbf{x} \in \mathcal{D}^d$, this operation is defined as:
\begin{equation}
\mathbf{M} \otimes_c \mathbf{x} = \frac{1}{\sqrt{c}} \tanh\left(\frac{\|\mathbf{M}\mathbf{x}\|}{\|\mathbf{x}\|} \tanh^{-1}(\sqrt{c}\|\mathbf{x}\|)\right) \frac{\mathbf{M}\mathbf{x}}{\|\mathbf{M}\mathbf{x}\|},
\end{equation}
where $\mathbf{M}\mathbf{x}$ represents standard matrix-vector multiplication. This operation preserves hyperbolic geometry while enabling learnable transformations. The key insight is that appropriate scaling matrices can systematically reduce embedding radii. By learning these matrices during training, our framework can make fine-grained information accessible for graph learning tasks while adapting to the specific requirements of different datasets and task types.

\subsection{Hierarchical Contrastive Learning}
We optimize the radius-adjusted embeddings through a contrastive learning objective designed for hyperbolic space. This objective encourages alignment between corresponding graph and text representations while leveraging the hierarchical structure encoded by radial positions. The training process operates on batches of graph-text pairs. Positive pairs correspond to the same semantic content. Negative pairs represent different concepts. 

For a batch $\{(\mathbf{G}_i, \mathbf{T}_i)\}_{i=1}^B$, we define the hyperbolic contrastive loss as:
\begin{equation}
\mathcal{L}_{\text{align}} = -\frac{1}{B} \sum_{i=1}^B \log \frac{\exp(-d_c(\tilde{\mathbf{z}}_{g,i}, \tilde{\mathbf{z}}_{t,i})/\tau)}{\sum_{j=1}^B \exp(-d_c(\tilde{\mathbf{z}}_{g,i}, \tilde{\mathbf{z}}_{t,j})/\tau)},
\end{equation}
where $d_c(\mathbf{x}, \mathbf{y})$ represents hyperbolic distance between points $\mathbf{x}$ and $\mathbf{y}$. The temperature parameter $\tau > 0$ controls distribution sharpness. The hyperbolic distance is computed as:
\begin{equation}
d_c(\mathbf{x}, \mathbf{y}) = \frac{2}{\sqrt{c}} \tanh^{-1}(\sqrt{c}\|\ominus_c \mathbf{x} \oplus_c \mathbf{y}\|)
\end{equation}
This distance metric naturally incorporates hierarchical relationships. Embeddings at similar radial levels are more compatible than those at different abstraction levels. This property helps our framework learn appropriate radius adjustments that enhance both alignment quality and representation expressiveness. To prevent degenerate solutions where embeddings collapse to the origin, we introduce a regularization term:
\begin{equation}
\mathcal{L}_{\text{reg}} = \lambda_r \sum_{k=1}^K \left( \|\mathbf{S}_{g,k} - \mathbf{I}_n\|_F^2 + \|\mathbf{S}_{t,k} - \mathbf{I}_n\|_F^2 \right),
\end{equation}
where $\mathbf{I}_n$ is the $n \times n$ identity matrix and $\|\cdot\|_F$ denotes Frobenius norm. The regularization strength $\lambda_r > 0$ encourages scaling matrices to remain close to identity when no adjustment is needed. This prevents unnecessary distortion of the embedding space while maintaining training stability. The regularization also helps the model learn interpretable radius adjustments that reflect genuine task requirements rather than arbitrary transformations. The complete training objective combines both components:
\begin{equation}
\mathcal{L}_{\text{total}} = \mathcal{L}_{\text{align}} + \mathcal{L}_{\text{reg}}
\end{equation}
This formulation enables the model to learn optimal radius adjustments that enhance graph-text alignment while maintaining robust hyperbolic representations. The learned adjustments reflect the balance between accessing fine-grained information and preventing embedding collapse.

\subsection{Zero-Shot Inference}
During inference on target graphs, H4G applies learned radius adjustments to new graph-text pairs without additional fine-tuning. This enables faithful zero-shot transfer by preserving fine-grained structural patterns. The capability is important for practical applications where labeled target data is unavailable. Given a target graph $\mathcal{G}^{\text{target}}$ with nodes $\{v_1, v_2, \ldots, v_M\}$ and class descriptions $\{\text{class}_1, \text{class}_2, \ldots, \text{class}_C\}$, we first encode both components using trained encoders. Graph nodes are processed through the graph neural network to obtain node representations. These are then projected to hyperbolic space and adjusted using learned scaling matrices $\mathbf{S}_g$. Similarly, class descriptions are encoded through the text encoder, projected to hyperbolic space, and adjusted using $\mathbf{S}_t$. This parallel processing ensures both graph and text representations undergo consistent radius adjustment. Node predictions identify the class with minimum hyperbolic distance to the adjusted node embedding:
\begin{equation}
\hat{y}_j = \arg\min_{k \in \{1,\ldots,C\}} d_c(\tilde{\mathbf{z}}_{g,j}^{\text{target}}, \tilde{\mathbf{z}}_{t,k}^{\text{class}}),
\end{equation}
where $\tilde{\mathbf{z}}_{g,j}^{\text{target}}$ represents the radius-adjusted embedding of node $j$ in the target graph. $\tilde{\mathbf{z}}_{t,k}^{\text{class}}$ denotes the radius-adjusted embedding of class $k$. This approach enables effective zero-shot transfer by leveraging fine-grained representations learned through systematic radius reduction. The learned radius adjustments generalize across different graph domains, allowing the model to maintain consistent performance without domain-specific tuning.
\begin{table}[h]
    \centering
    \caption{Statistics for node-classification and node-description datasets.}
    \label{table1}
    \setlength{\tabcolsep}{3.2mm}{
    \begin{tabular}{lrrc}
        \toprule
        Dataset & \#Nodes     & \#Edges & Avg. Degree \\
        \midrule
        ogbn-arxiv & 169,343 & 1,166,243 & 13.7 \\
        ogbn-products & 2,449,029 & 61,859,140 & 50.5 \\
        cora & 2,708 & 5,429 & 4.0 \\
        pubmed & 19,717 & 44,338 & 4.5 \\
        Children & 76,875 & 1,554,578 & 40.4 \\
        History & 41,551 & 358,574 & 17.2 \\
        Computers & 87,229 & 721,081 & 16.5 \\
        Photo & 48,362 & 500,928 & 20.7  \\
        \bottomrule
    \end{tabular}
    }
    \normalsize
    \vspace*{-10pt}
\end{table}

\begin{table*}[ht]
\centering
\captionsetup{skip=6pt}
\caption{The accuracy of different shot node classification on two OGB datasets. }
\label{table2}
\setlength{\tabcolsep}{0.9mm}{
\begin{tabular}{l|cccccccc} 
\toprule
            Dataset    & \multicolumn{4}{c}{ogbn-arxiv} & \multicolumn{4}{c}{ogbn-products} \\
\cmidrule(lr){1-1}\cmidrule(lr){2-5}\cmidrule(lr){6-9} 
            Setting    & 5-way 0-shot & 5-way 5-shot & 10-way 0-shot & 10-way 5-shot & 5-way 0-shot & 5-way 5-shot & 10-way 0-shot & 10-way 5-shot \\
\midrule
OGB features  & 48.98±0.77 & 59.31±0.71 & 35.63±0.85 & 44.59±0.61 & 40.42±1.30 & 50.57±1.83 & 30.17±0.88 & 38.75±0.61\\
GPN    & 56.71±0.67 & 66.34±0.78 & 43.59±0.25 & 53.60±0.86 & 58.92±1.92 & 69.86±1.90 & 48.47±0.66 & 58.36±0.45\\
G-Meta & 55.26±1.56  & 63.77±2.53 & 44.49±0.94 & 52.78±1.04 & 58.94±1.18 & 66.02±0.71 & 47.08±1.48 & 57.81±1.40\\
TENT   & 57.26±0.35  & 65.80±0.37 & 41.48±0.54 & 51.84±0.86 & 57.68±2.78 & 66.63±1.48 & 43.82±0.59	& 51.53±0.90\\
\midrule
GIANT  & 67.62±1.59 & 78.18±1.25 & 55.25±1.20 & 64.96±0.69 & 63.88±1.40 & 75.46±1.53 & 54.50±0.64 & 65.67±0.38\\
LLaGA  & 69.75±1.43 & 80.56±1.67 & 56.79±0.89 & 67.14±1.12 & 65.58±1.86 & 77.93±1.24 & 55.62±0.73 & 67.89±0.92\\
GraphEdit & \underline{70.78±1.21} & 79.87±1.38 & 57.88±1.45 & 66.38±0.97 & 65.00±2.03 & 76.84±1.95 & 56.79±1.16 & 66.94±0.81\\
TEA-GLM & 68.71±1.76 & \underline{81.02±1.54} & 56.38±1.03 & \underline{67.85±0.86} & \underline{67.08±1.52} & 77.18±1.41 & \underline{57.39±0.95} & 68.52±1.23\\
GraphCLIP & 70.09±1.92 & 80.25±1.81 & \underline{58.39±1.28} & 66.72±1.35 & 65.99±1.67 & \underline{78.35±1.78} & 56.07±1.08 & \underline{69.16±0.76}\\
GraphTranslator & 69.36±1.35 & 79.43±1.19 & 57.19±0.92 & 67.41±0.78 & 66.29±1.94 & 76.52±1.62 & 56.38±0.87 & 67.35±1.14\\
\midrule
\rowcolor{gray!20}
\textbf{H4G (Ours)} & \textbf{80.77±1.28} & \textbf{86.74±1.45} & \textbf{68.51±1.38} & \textbf{75.21±1.02} & \textbf{78.46±1.41} & \textbf{84.13±1.69} & \textbf{65.33±0.98} & \textbf{72.48±0.56}\\
\bottomrule
\end{tabular}
}
\end{table*}

\begin{table*}[!ht]
\centering
\captionsetup{skip=6pt}
\caption{The accuracy of different shot node classification on four Amazon Review datasets.}
\label{table3}
\begin{minipage}{\textwidth}
\centering
\setlength{\tabcolsep}{3.8mm}{
\begin{tabular}{l|cccccc} 
\toprule
            Dataset & \multicolumn{3}{c}{Children} & \multicolumn{3}{c}{History} \\
\cmidrule(lr){1-1}\cmidrule(lr){2-4}\cmidrule(lr){5-7} 
            Setting & 3-way 0-shot & 3-way 5-shot & 3-way 10-shot & 3-way 0-shot & 3-way 5-shot & 3-way 10-shot \\
\midrule
node features & 36.50±0.61 & 43.33±1.78 & 49.00±0.77 & 34.08±1.63 & 38.16±1.33 & 40.82±1.07 \\
GPN  & 50.39±0.42 & 60.51±1.16 & 63.89±0.93 & 35.40±0.74 & 40.63±0.82 & 43.88±1.01 \\
G-Meta & 49.68±2.38 & 57.76±1.43 & 61.62±1.52 & 37.18±1.32 & 41.11±0.76 & 42.50±0.86 \\
TENT   & 48.97±0.76 & 60.52±1.73 & 64.32±0.81 & 35.24±0.72 & 37.73±0.43 & 41.47±2.05 \\
LLaGA  & 52.27±1.14 & 62.38±0.92 & 66.15±1.43 & 39.34±1.58 & 43.29±1.21 & 45.93±0.68 \\
GraphEdit & 53.49±0.87 & 63.71±1.55 & 65.48±0.76 & 38.20±0.93 &  \underline{46.73±1.39}& 46.71±1.15 \\
TEA-GLM & 52.71±1.62 &  64.82±0.71& 67.23±1.29 & 39.73±1.41 & 44.52±0.85 & \underline{49.62±0.91} \\
GraphCLIP & 54.71±1.31 & \underline{65.16±1.24} & 68.91±0.94 & 41.06±0.67 & 45.16±1.02 & 48.25±1.27 \\
GraphTranslator & \underline{55.42±0.95} & 61.94±1.08 & \underline{69.57±1.18} & \underline{42.24±1.85} & 42.87±1.74 & 47.38±1.93 \\
\midrule
\rowcolor{gray!20}
\textbf{H4G (Ours)}  & \textbf{76.28±1.13} & \textbf{72.43±1.46} & \textbf{75.53±1.27} & \textbf{61.97±2.10} & \textbf{57.36±1.28} & \textbf{54.81±0.47} \\
\bottomrule
\end{tabular} 
}
\end{minipage}

\vspace{1pt}

\begin{minipage}{\textwidth}
\centering
\setlength{\tabcolsep}{3.8mm}{
\begin{tabular}{l|cccccc} 
\toprule
            Dataset   & \multicolumn{3}{c}{Computers} & \multicolumn{3}{c}{Photo} \\
\cmidrule(lr){1-1}\cmidrule(lr){2-4}\cmidrule(lr){5-7} 
            Setting    & 3-way 0-shot & 3-way 5-shot & 3-way 10-shot & 3-way 0-shot & 3-way 5-shot & 3-way 10-shot \\
\midrule
node features & 34.22±1.36 & 40.11±1.72 & 44.55±1.11 & 37.20±0.64 & 44.41±1.68 & 50.57±1.38\\
GPN  & 65.99±1.67 & 70.47±1.72 & 71.56±1.91 & 68.76±1.75 & 76.01±1.12 & 70.77±0.97\\
G-Meta & 65.85±2.19 & 71.14±2.52 & 72.36±0.63 & 64.11±1.08 & 72.41±1.09 & 73.54±1.22\\
TENT   & 55.83±1.94 & 62.13±1.17 & 65.96±1.92 & 62.78±1.20 & 71.20±1.30 & 72.19±2.03\\
LLaGA  & 67.02±0.73 & 73.27±1.85 & 74.63±1.04 & 69.85±1.67 & 77.48±0.84 & 74.21±1.45\\
GraphEdit & \underline{68.23±1.42} & 72.89±0.96 & 73.91±1.71 & \underline{70.67±1.02} & \underline{78.63±1.28} & 75.86±0.73\\
TEA-GLM & 67.55±1.18 & \underline{74.65±1.53} & \underline{75.28±0.89} & 70.05±1.91 & 77.92±1.56 & 76.43±1.19\\
GraphCLIP & 67.13±2.05 & 73.78±1.21 & 74.82±1.48 & 69.34±0.89 & 76.85±1.74 & \underline{77.62±0.86}\\
GraphTranslator & 67.79±0.91 & 74.32±1.67 & 75.14±1.32 & 70.46±1.35 & 78.21±0.97 & 76.95±1.61\\
\midrule
\rowcolor{gray!20}
\textbf{H4G (Ours)}  & \textbf{74.43±2.16} & \textbf{76.17±0.54} & \textbf{79.77±1.06} & \textbf{76.14±1.77} & \textbf{80.68±0.99} & \textbf{82.89±0.94} \\
\bottomrule
\end{tabular} 
}
\end{minipage}
\vspace{-5pt}
\end{table*}

\begin{table*}[ht]
\centering
\captionsetup{skip=6pt}
\caption{The Sbert Score of different shot node description on Cora and Pubmed datasets.}
\label{table4}
\setlength{\tabcolsep}{0.9mm}{
\begin{tabular}{l|cccccccc} 
\toprule
Dataset    & \multicolumn{4}{c}{Cora} & \multicolumn{4}{c}{Pubmed} \\
\cmidrule(lr){1-1}\cmidrule(lr){2-5}\cmidrule(lr){6-9} 
Setting    & 5-way 0-shot & 5-way 5-shot & 10-way 0-shot & 10-way 5-shot & 5-way 0-shot & 5-way 5-shot & 10-way 0-shot & 10-way 5-shot \\
\midrule
GPN        & 38.03±0.67 & 42.18±0.78 & 39.79±0.85 & 44.73±0.86 & 43.07±1.92 & 47.34±1.90 & 44.31±0.66 & 48.93±0.45\\
G-Meta     & 46.44±1.56 & 51.53±2.53 & 48.64±0.94 & 53.95±1.04 & 51.46±1.18 & 56.68±0.71 & 52.92±1.48 & 58.27±1.40\\
TENT       & 47.98±0.35 & 53.02±0.37 & 49.85±0.54 & 55.26±0.86 & 52.76±2.78 & 58.12±1.48 & 54.31±0.59 & 59.65±0.90\\
\midrule
GIANT      & 59.54±1.59 & 65.38±1.25 & 61.13±1.20 & 67.84±0.69 & 73.20±1.40 & 80.23±1.53 & 74.87±0.64 & 82.14±0.38\\
LLaGA      & 62.05±1.44 & 68.73±1.18 & \underline{64.51±0.82} & 69.86±0.75 & 76.11±1.23 & \underline{84.15±1.67} & 78.13±0.88 & 85.38±0.56\\
GraphEdit  & \underline{62.75±1.21} & 67.96±1.35 & 63.83±0.96 & \underline{71.25±0.91} & \underline{76.49±1.54} & 83.47±1.41 & \underline{78.44±0.73} & 84.71±0.82\\
TEA-GLM    & 63.48±1.37 & 70.00±1.42 & 65.32±1.15 & 72.00±0.88 & 77.28±1.62 & 85.00±1.74 & 79.12±0.95 & 86.50±0.71\\
GraphCLIP  & 61.58±1.28 & \underline{69.24±1.09} & 63.26±0.77 & 70.54±0.83 & 75.30±1.46 & 82.96±1.58 & 76.98±0.69 & \underline{85.92±0.47}\\
GraphTranslator & 62.60±1.52 & 68.52±1.24 & 64.22±0.89 & 70.93±0.72 & 75.79±1.31 & 83.71±1.45 & 77.72±0.81 & 84.26±0.63\\
\midrule
\rowcolor{gray!20}
\textbf{H4G (Ours)}        & \textbf{70.66±1.18} & \textbf{73.65±1.34} & \textbf{72.67±1.26} & \textbf{75.78±0.94} & \textbf{85.71±1.27} & \textbf{89.42±1.55} & \textbf{87.79±0.91} & \textbf{91.08±0.49}\\
\bottomrule
\end{tabular}
}
\end{table*}

\begin{table}[tbp] 
    \centering
    \small
     \addtolength{\tabcolsep}{-0.5mm}
    \caption{Data ablation study with an increasing number of source domains, while fixing \emph{Cora} as the target domain.
    }
    \label{table5}%
    \resizebox{1\linewidth}{!}{%
    \begin{tabular}{l|cccc}
    \toprule
   \multirow{2}*{Method} &\multicolumn{4}{c}{Number of source domains}\\   & 1 & 2 & 3 & 4
      \\\midrule
    GraphEdit
    & 36.24\text{\scriptsize ±12.06}
    & 37.87\text{\scriptsize ±11.79}
    & 37.82\text{\scriptsize ±11.23}
    & 39.70\text{\scriptsize ±11.84}  
\\
    GraphCLIP
    & 40.26\text{\scriptsize ±12.14}
    & 37.36\text{\scriptsize ±\phantom{0}9.46}
    & 36.16\text{\scriptsize ±11.99}
    & 39.77\text{\scriptsize ±12.74}
 \\
      H4G
    & \textbf{42.05}\text{\scriptsize ±11.00}
    & \textbf{43.65}\text{\scriptsize ±11.01}
    & \textbf{44.20}\text{\scriptsize ±11.56}
    & \textbf{48.25}\text{\scriptsize ±12.96}
\\    \bottomrule
        \end{tabular}}
\end{table}
\begin{table}[tb]
    \centering
    \small
    \addtolength{\tabcolsep}{-.1mm}
    \caption{Analysis of cross-domain node classification with different source-target domain combinations.}
    \label{table6}%
    \begin{tabular}{@{}c|c|ccc@{}}
    \toprule
    Target & \multirow{2}*{Source domains} & \multicolumn{3}{c}{Accuracy (\%)} \\
    domain &  & LLaGA & TEA-GLM & H4G  \\
    \midrule
    Products
    & ArXiv, PubMed, Cora &  
    13.8 & 9.1 & \textbf{18.5}\text{\scriptsize (4.7↑)} \\
    PubMed
    & ArXiv, Cora &
    42.1  & 38.6 & \textbf{49.6}\text{\scriptsize (11.0↑)}\\
    ArXiv
    & ArXiv, Cora &  
    48.2  & 50.3 & \textbf{58.1}\text{\scriptsize (7.8↑)}\\   
    Cora
    & ArXiv, PubMed &  
    16.4  & 21.9 & \textbf{46.4}\text{\scriptsize (4.5↑)}\\
    ArXiv
    & ArXiv, PubMed &  
    46.7  & 54.7 & \textbf{71.3}\text{\scriptsize (7.4↑)}\\
    \bottomrule
    \end{tabular}\\
       \parbox{1\linewidth}{\footnotesize The numbers in parentheses indicate the performance improvement of H4G over the second-best baseline.}
\end{table}

\begin{figure*}[t]
   \centering
   \includegraphics[width=\textwidth]{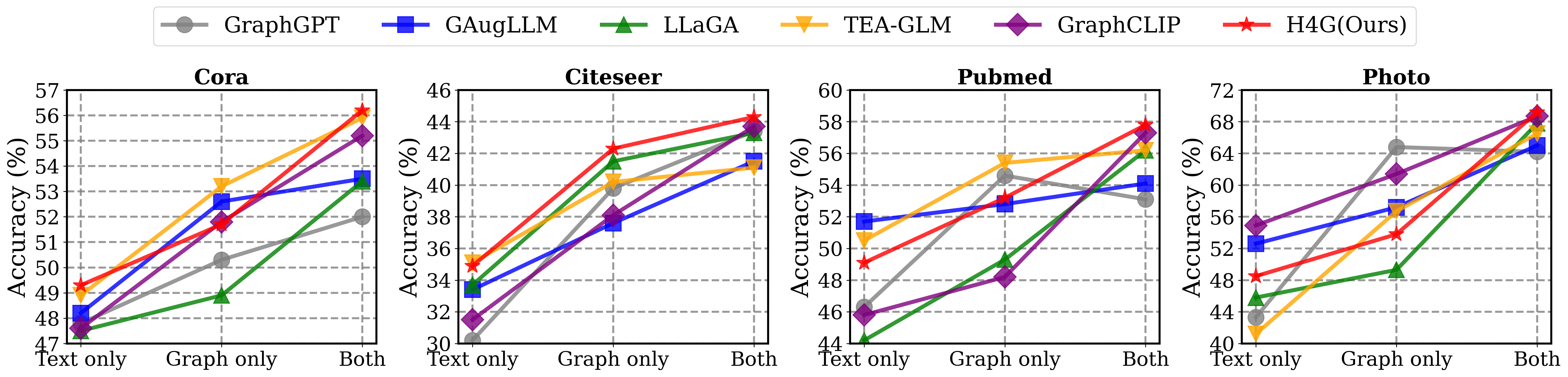}
\caption{Ablation study on hyperbolic projection strategies across four datasets. "Text only" applies hyperbolic projection solely to text embeddings while keeping graph embeddings in Euclidean space. "Graph only" projects only graph representations into hyperbolic space. "Both" represents our full H4G framework with dual hyperbolic projection. Results demonstrate that projecting both modalities into hyperbolic space achieves optimal alignment quality and superior zero-shot transfer performance.}
   \label{figure4}
\end{figure*}

\section{Experiments}

\subsection{Datasets}
We evaluate H4G on 8 text-attributed graph datasets to assess its radius reduction effectiveness \cite{hu2020open}. Following zero-shot graph learning conventions, we divide datasets into source and target sets for pretraining and evaluation respectively. Source datasets include ogbn-ArXiv, PubMed \cite{sen2008collective}, and ogbn-Products from academic citation networks and e-commerce platforms, providing diverse graph structures and rich textual information for robust training. Target datasets include Cora  \cite{yang2016revisiting} from citation networks, and Children \cite{mcauley2015image,chiang2019cluster}, History, Computers, and Photo from e-commerce product networks, enabling transfer evaluation across varying domains, sizes, and structural properties. Each dataset contains node-level textual descriptions ranging from paper abstracts to product information, naturally aligning with graph structures for hyperbolic learning. This setup comprehensively evaluates how radius reduction facilitates fine-grained representation transfer across diverse graph types.

\subsection{Baselines}
We compare H4G with nine state-of-the-art methods spanning three categories. The first category includes traditional GNN-based approaches adapted for zero-shot learning such as GPN \cite{ding2020graph}, G-Meta \cite{huang2020graph}, and TENT \cite{wang2021tent}, which leverage self-supervised objectives and metric learning for cross-dataset transfer. The second category includes graph-text alignment methods including GIANT for structural-semantic bridging, GraphCLIP \cite{graphclip} for contrastive alignments in Euclidean space, and GraphTranslator \cite{graphtranslator} for prompt-based cross-domain transfer. The third category includes LLM-enhanced approaches like LLaGA \cite{chen2024llaga} that encodes graph structures as token sequences, GraphEdit \cite{GraphEdit} for graph-conditioned text generation, and TEA-GLM \cite{wang2024llms} that integrates textual and structural information through language models. These baselines range from pure graph learning to sophisticated text-graph integration, enabling comprehensive evaluation of systematic radius reduction in hyperbolic space.

\subsection{Evaluation Metrics and Models}
We evaluate H4G under zero-shot and few-shot settings using standard graph tasks. For zero-shot, we assess node classification and link prediction on target datasets without additional training and report mean accuracy with standard deviation and AUC scores. For few-shot, we test 0, 3, 5, and 10 shots per class to examine how reduced radii facilitate adaptation. The graph encoder uses a 12-layer GraphGPS architecture with 1024 hidden dimensions, while the text encoder is a frozen SBERT model with matching 384-dimensional hyperbolic space. Through systematic radius reduction, H4G consistently achieves better fine-grained information preservation and demonstrates stable scaling factors across diverse datasets, confirming radius reduction as a robust approach for accessing critical graph structures.

\begin{table*}[htp]
    \centering
    \caption{Ablation study of different components in H4G on zero-shot node classification.}
    \label{table7}
    \vspace{-1em}
    \resizebox{\textwidth}{!}{
    \begin{tabular}{l|cccccccc}
    \toprule
    Methods & ogbn-arxiv & ogbn-products & Cora & PubMed & Children & History & Computers & Photo \\ 
    \midrule
    H4G (Full) & \textbf{80.77±1.28} & \textbf{78.46±1.41} & \textbf{34.13±1.85} & \textbf{91.56±0.82} & \textbf{76.28±1.13} & \textbf{61.97±2.10} & \textbf{74.43±2.16} & \textbf{76.14±1.77} \\
    \quad w/o Radius Adjustment & 72.28±1.54 & 70.15±1.69 & 28.75±2.03 & 84.42±1.12 & 68.31±1.47 & 54.18±2.38 & 66.59±2.41 & 68.45±1.95 \\
    \quad w/o Hyperbolic Space & 75.61±1.43 & 73.24±1.57 & 30.89±1.94 & 87.68±0.96 & 71.76±1.32 & 57.08±2.22 & 69.78±2.28 & 71.29±1.87 \\
    \quad w/o Block-Diagonal & 77.15±1.37 & 75.48±1.50 & 31.84±1.89 & 89.23±0.89 & 73.49±1.24 & 59.35±2.15 & 71.61±2.21 & 73.19±1.83 \\
    \quad w/o Regularization & 78.89±1.32 & 76.79±1.46 & 32.91±1.87 & 90.34±0.85 & 74.84±1.19 & 60.28±2.12 & 72.82±2.18 & 74.51±1.80 \\
    \bottomrule
    \end{tabular}
    }
    \vspace{-1em}
\end{table*}
\begin{figure*}[t]
   \centering
   \includegraphics[width=\textwidth]{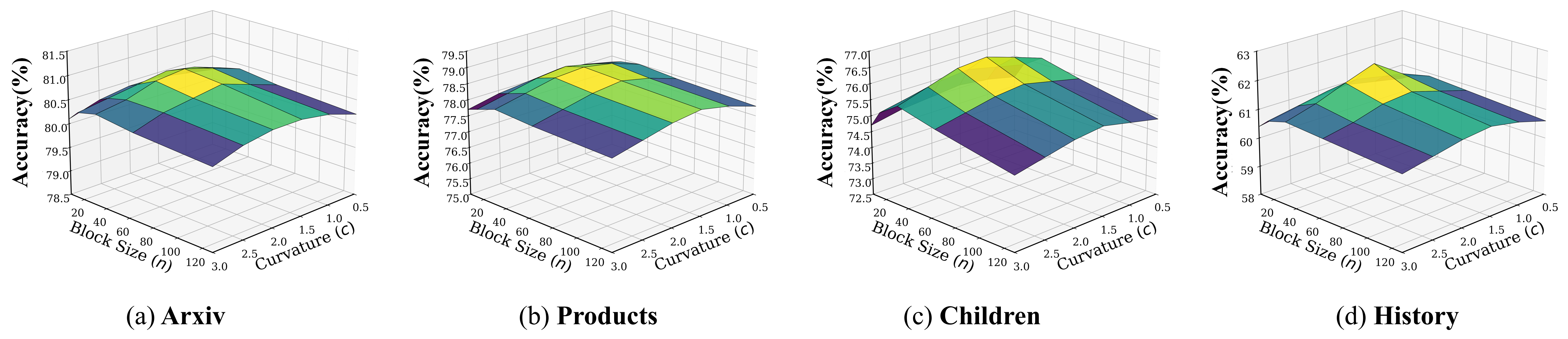}
\caption{Sensitivity analysis of block size and curvature on zero-shot node classification accuracy. The surface demonstrates that H4G achieves optimal performance when block size equals 32 and curvature equals 1.0, validating the effectiveness of fine-grained radius control in appropriately curved hyperbolic space. Performance degrades when block size becomes too small or too large, or when curvature deviates significantly from the optimal value.}
   \label{figure5}
\end{figure*}

\subsection{Implementation Details}
H4G is implemented in PyTorch with hyperbolic operations using geoopt for stable Poincaré ball computations. We use AdamW optimizer with learning rates of 1e-4 for encoders and 5e-5 for scaling matrices. Training runs for 100 epochs with batch size 256 and gradient clipping (max norm 1.0). The curvature parameter is set to $c = 1.0$, and block size $n = 32$ results in $K = d/32$ blocks, adding only 4.8\% parameters. Scaling matrices $\mathbf{S}_g$ and $\mathbf{S}_t$ initialize near identity ($\mathbf{I}_n + \epsilon$, $\epsilon \sim \mathcal{N}(0, 0.01)$) for gradual radius reduction. We set temperature $\tau = 0.07$ and regularization strength $\lambda_r = 0.01$. The graph encoder uses 12-layer GraphGPS with 1024 hidden dimensions, while the text encoder uses frozen SBERT with 384-dimensional embeddings projected to hyperbolic space. All experiments run on NVIDIA A100 80GB SXM4 GPUs, with pretraining taking approximately 12 hours. Learned scaling factors consistently converge between 0.3 and 0.7, reducing embedding radii from 7-8 to 3-4, validating that radius reduction enables access to fine-grained structural information for effective zero-shot transfer.

\subsection{Main Result}

Tables~\ref{table2} and~\ref{table3} demonstrate H4G's superior zero-shot performance across all datasets. On OGB benchmarks, H4G achieves 80.77\% on ogbn-arxiv and 78.46\% on ogbn-products, surpassing GraphCLIP by 10.68\% and 12.47\% respectively. The improvements are even more pronounced on heterophilic Amazon graphs, with gains of 21.57\% on Children and 7.30\% on Computers. These consistent improvements validate that systematic radius reduction enables faithful preservation of fine-grained structural patterns essential for effective zero-shot transfer.

\subsection{Model Analysis}

Figure~\ref{figure4} examines the impact of applying hyperbolic projection to different modalities. The "Both" configuration consistently outperforms single-modality projections by 4-8\% across all datasets. Notably, "Graph only" surpasses "Text only", indicating that structural hierarchies benefit more from hyperbolic geometry. Even single-modality projection provides 2-5\% gains over Euclidean baselines, confirming that unified geometric representation in hyperbolic space is crucial for optimal graph-text alignment.

\subsection{Hyper-parameter Analysis}

Figure~\ref{figure5} visualizes hyperparameter sensitivity through 3D surfaces on four datasets. Ogbn-arxiv exhibits a broad plateau with minimal fluctuations, while Children shows a pronounced ridge along block size 32, demonstrating that heterophilic graphs require precise granularity control. Despite varying sensitivity patterns, all datasets consistently achieve optimal performance around $c=1.0$ and $n=32$, validating our design choices and demonstrating reasonable robustness within practical parameter ranges.

\subsection{Ablation Study}

Table~\ref{table5} quantifies each component's contribution. Removing radius adjustment causes the largest drop (8.49\% on arxiv, 8.28\% on products), confirming it as the core innovation. Ablating hyperbolic space results in 5.09\% and 5.22\% declines, while removing block-diagonal structure and regularization lead to smaller drops of 3.56\% and 1.83\% respectively. Tables~\ref{table6} and~\ref{table7} further demonstrate H4G's scalability with increasing source domains and superior cross-domain transfer, confirming all components contribute positively with radius adjustment as the fundamental mechanism.

\section{Conclusion}
We address zero-shot graph learning's over-abstraction issue by optimizing embedding radii in hyperbolic space. Large radii compress critical structural details, motivating H4G, which restores multi-scale information via learnable radius reduction. Experiments show consistent improvements, while the radius-granularity relationship offers a principled way to tackle information loss. Findings suggest explicit abstraction control enhances transferability, paving the way for adaptive radius adjustment and multi-scale representation advancements.

\bibliographystyle{ACM-Reference-Format}
\bibliography{sample-base}

\end{document}